# LLM-DER:A Named Entity Recognition Method Based on Large Language Models for Chinese Coal Chemical Domain

## LE XIAO*

School of Information Science and Engineering, Henan University of Technology,Zhengzhou,China,xiaole@haut.edu.cn

## YUNFEI XU

School of Information Science and Engineering, Henan University of Technology,Zhengzhou,China,xuyunfei@stu.haut.edu.cn

## JING ZHAO

School of Information Science and Engineering, Henan University of Technology,Zhengzhou,China,Zjing@stu.haut.edu.cn

Domain-specific Named Entity Recognition (NER), whose goal is to recognize domain-specific entities and their categories, provides an important support for constructing domain knowledge graphs. Currently, deep learning-based methods are widely used and effective in NER tasks, but due to the reliance on large-scale labeled data. As a result, the scarcity of labeled data in a specific domain will limit its application.Therefore, many researches started to introduce few-shot methods and achieved some results. However, the entity structures in specific domains are often complex, and the current few-shot methods are difficult to adapt to NER tasks with complex features.Taking the Chinese coal chemical industry domain as an example,there exists a complex structure of multiple entities sharing a single entity, as well as multiple relationships for the same pair of entities, which affects the NER task under the sample less condition.In this paper, we propose a Large Language Models (LLMs)-based entity recognition framework LLM-DER for the domain-specific entity recognition problem in Chinese, which enriches the entity information by generating a list of relationships containing entity types through LLMs, and designing a plausibility and consistency evaluation method to remove misrecognized entities, which can effectively solve the complex structural entity recognition problem in a specific domain.The experimental results of this paper on the Resume dataset and the self-constructed coal chemical dataset Coal show that LLM-DER performs outstandingly in domain-specific entity recognition, not only outperforming the existing GPT-3.5-turbo baseline, but also exceeding the fully-supervised baseline, verifying its effectiveness in entity recognition.

# 1 INTRODUCTION

The goal of Named Entity Recognition (NER) is to identify mentioned entities in a sentence and classify them into predefined entity types. In general domains [2], these entity types can include names of people, places, organizations, time, etc. Whereas in specific domains, further identification of entity types specific to that domain is required. For example, in the biomedical domain [3], entities such as diseases and chemical substances need to be identified.Currently, deep learning-based methods are widely used in NER tasks [4], and advanced results have been achieved [5,6]. Compared to the traditional method of manually constructing features, deep learning-based methods do not need to manually design features and can automatically obtain models by training a large amount of data. Among them, based on the excellent sequence modeling capability of unidirectional Long Short-Term Memory (LSTM) models, many methods use LSTM Conditional Random Fields (CRF) as the main framework for the NER task and incorporate a variety of related features based on it [7].BiLSTM-CRF is the most commonly used method, with which state-of-the-art performance is achieved using this method as the main framework[4].However, deep learning models rely on a large amount of labeled data for training [8,9] in order to efficiently extract entity features, and the scarcity of labeled data may limit the performance of deep learning methods in specific domains because large-scale labeled data is very expensive and time-consuming. Although few-shot learning methods can alleviate the data scarcity problem to a certain extent, it is difficult to achieve good results in entity recognition in Chinese specific domains with complex structures [1]. For example, the field of coal chemical industry covers numerous products with complex upstream and downstream relationships. A product can usually be obtained through multiple synthesis pathways, and in the process of preparing downstream products, it often needs to rely on multiple upstream chemical raw materials. In addition, different chemical products may substitute each other in application scenarios. We found that entities within the Chinese coal chemical industry usually consist of multiple vocabularies through their data analysis and experimental comparisons(as shown in Figure 1), and there are complex situations in which multiple entities share a single entity, as well as the same pair of entities has multiple relationships, leading to the inability of the few-shot methods to adapt to the NER task with complex structural features.

However, recent studies [36,37] have shown that large language models (LLMs) such as GPT-3 and ChatGPT, even without any training or fine-tuning, rely solely on in-context learning (ICL) methods to make predictions by concatenating a query with a few-shot demonstrations to prompt LLMs. It also performs well in various natural language processing (NLP) downstream tasks [11]. Because LLMs is better able to capture contextual information in texts and infer the boundaries and types of entities through context, which helps to identify entities with complex and specialized structures in domain-specific texts. And LLMs can perform transfer learning, where knowledge learned from texts in other domains helps to process entities in texts from different domains [18].To this end, the LLM-DER framework is proposed, which enhances entity contextual relevance by adding entity type information at both ends of predefined relationships and using this information to generate a list of similar relationships, and removes misrecognized entities through a plausibility and consistency weighted assessment. We evaluate our proposed method on the self-built Chinese coal chemical domain dataset Coal and the public dataset Resume. The experimental results show that LLM-DER can effectively extract domain-specific entity information and exceed the fully supervised baseline.

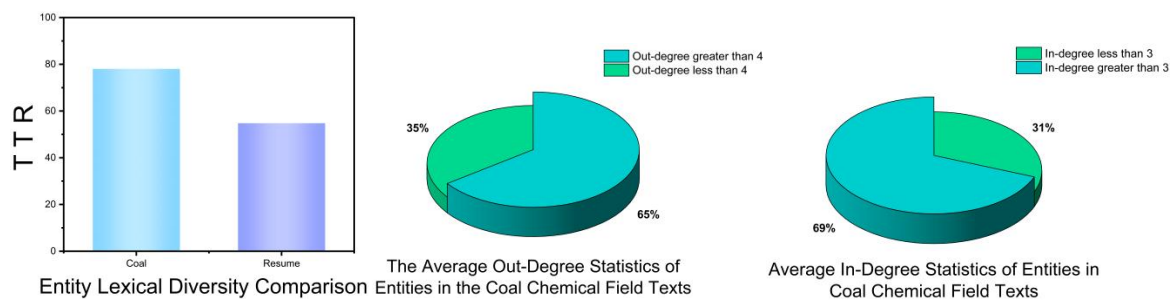

Figure 1.The first left shows the entity lexical diversity statistics, using Type Token Ratio (TTR) as a measure to reflect the lexical diversity. In the comparison, the coal chemical dataset "Coal" shows higher lexical diversity than the Chinese resume dataset "Resume" used as a control. The second and third from the left show the average out-degree and in-degree of entities in the coal-chemical domain dataset. The second from the left shows that the percentage of out-degrees exceeding the average value of 4 is 65%; the third from the left shows that the percentage of in-degrees exceeding the average value of 3 is 69%.

Our work has contributed to the following:

●A domain-specific NER approach is proposed, which first strengthens the relevance of entities in context by adding entity type information to both ends of a predefined relationship. Subsequently, a list of relationships is generated using LLMs to perform entity recognition in a relationship-driven manner.

●A domain-specific NER framework LLM-DER based on LLMs is constructed, which is designed with consistency and plausibility weighted assessment modules to effectively remove misidentified entities.

●The effectiveness of the present entity recognition method is verified on Resume and Coal Chemical datasets, which provides a new means of entity extraction for the next step of constructing industrial maps.

## 2 RELATED WORK

### 2.1 Named Entity Recognition

Named entity recognition methods mainly include: rule-based methods, statistical model-based methods and deep learning-based methods. Earlier, named entities were mainly recognized by manually writing rules, which required domain experts to manually construct recognition rules, which is time-consuming and laborious. Unlike rule-based approaches, statistical model-based approaches do not require domain experts to formulate rules and use Hidden Markov Model (HMM) [17], Maximum Entropy Model (MEM) [32] or Conditional Random Fields Model (CRF) [21] on manually labeled datasets, etc. can extract data features better, but in practical applications, the labeling of data and the design of features limit the development of this method.

### 2.2 Deep Learning Based Approach

With the development of deep learning techniques, it has rapidly become a research hotspot due to its excellent performance on named entity recognition tasks.Collobert et al [12] proposed a CNN-CRF model based on convolutional neural network (CNN) and CRF, which achieved excellent performance compared to previous models.Ma et al [13] and Chiu et al [14] introduced CNN-extracted character features in a word-level representation based on the BiLSTM-CRF framework to enhance the expressive power of the word-level representation.Dong et al [15] composed each character into a root sequence and utilized Long Short-Term Memory Network (LSTM) to obtain the root information of Chinese characters.Zhang et al [16] proposed Lattice-LSTM method, which replaces the traditional LSTM cells with lattice

LSTMs to cleverly encode Chinese characters and potential words matched with the lexicon. Based on the Lattice-LSTM, Li, Xiaonan et al [19] proposed the FLAT model by transforming the dot matrix structure into a planar structure consisting of spans. The model makes full use of the Lattice information and has good parallelization ability based on the powerful performance of Transformer and good positional coding design.

In practical applications, especially in specific domains, obtaining large amounts of labeled data is expensive and time-consuming. To address this problem, few-shot learning methods proposes a series of algorithms aimed at rapidly extracting knowledge from a limited number of labeled samples and adapting to new learning tasks [20]. One common approach is to utilize pre-trained language models that are pre-trained on large amounts of textual data and are able to capture rich linguistic features and knowledge that can be fine-tuned on small amounts of labeled data to accomplish specific tasks [30]. Other approaches include the use of meta-learning strategies to learn how to quickly adapt to new tasks by training models on multiple tasks, and the use of data augmentation techniques to expand the training set [22]. Despite the progress made by few-shot learning methods in dealing with few-sample problems, they still face a number of problems [25] . In particular, the generalization ability of the model is limited when dealing with data with complex structure.

### 2.3 Approaches Based On Large Language Modeling

LLMs has learned rich linguistic knowledge and contextual information through pre-training on huge scale text corpus, which enables LLMs to infer the boundaries and types of entities by understanding contextual cues of the text, and helps to recognize named entities in the text [10]. Currently,there are some researches based on LLMs to do NER, for example, Polak et al[23] proposed ChatExtract, which consists of a set of designed prompts that both recognize sentences with data, extract the data, and ensure the correctness of the data through a series of follow-up questions.Wang et al[24] proposed GPT-NER, which firstly converts the sequence labeling task into a generative task that can be easily accomplished by LLMs, and then proposes a self-verification strategy that effectively solves the "illusion" problem of LLMs by asking LLMs whether the entities they recognize belong to the labeled entity labels.The models and methods mentioned above are analyzed and summarized in detail in the following, as shown in Table 1.

Table 1. Summary of NER methodology

| | Method Keywords | Advantages | Limitations | Method |
|---|---|---|---|---|
| Deep Learning Based Approaches | CNN-CRF | Does not rely on any segmentation methods or task-specific functionality, making it simpler and more versatile | Computational cost constraints when dealing with complex scenario parsing tasks that require consideration of long-term dependencies | Collobert et al[12] |
| | CNN-extracted character features introduced into the word -level representation of the BiLSTM-CRF framework. | The model does not require feature engineering or data preprocessing and is suitable for a variety of sequence labeling tasks. | In few-shot environments, models may struggle to fully utilize rich feature representations, which in turn affects performance | Ma et al [13] and Chiu et al [14] |
| | Form each character into a root sequence and use LSTM to obtain the root information of Chinese characters | Bidirectional LSTM-CRF neural network utilizing both character-level and head er-level representations | If spelling and contextual features are removed, their performance is significantly reduced | Dong et al[15] |
| | Lattice-LSTM | Encoding of input characters and potential words in the dictionary. Better utilizat | In a few-shot environment, the model relies on a Lattice structure that struggles to captu | Zhang et al[16] |

| | | ion of word and word order information | re valid word sequence paths, thus affecting the accuracy of entity recognition | |
| | FLAT | Excellent parallelization with full use of Lattice information | Reliance on dictionaries and pre-trained word embeddings may not fully integrate and utilize domain-specific knowledge | Li, Xiaonan et al[19] |
| Based on a large language modeling approach | ChatExtract | Fully automated and accurate data extraction from research papers with minimal background knowledge | The complexity of text containing multiple values can lead to extraction errors or phantom data | Polak et al[23] |
| | GPT-NER | Introducing a Self-Verification Strategy to Address the Illusion of LLMs | Input sentences are long, which may make it difficult to process long input sequences | Wang et al[24] |

To summarize, LLMs obtains strong comprehension and generation capabilities by pre-training on large-scale text corpora. This makes LLMs-based entity recognition a better solution to the deep learning less sample dilemma, but underperforms on specific domains with complex structures. Therefore, this paper addresses domain-specific complex structure entity recognition.

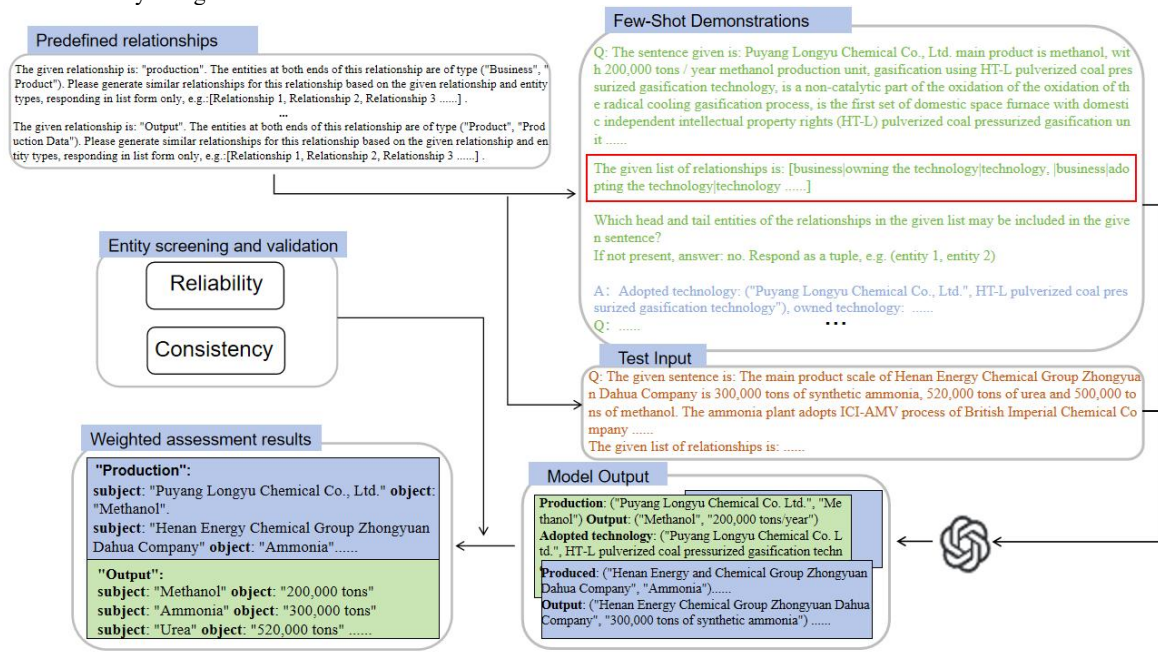

Figure 2: The overall framework diagram of LLM-DER, predefined relationships indicates that the LLMs is allowed to generate a list of relationships based on the command prompts, few-shot demonstrations indicates that the in-context learning capability of the LLMs is utilized to better the output results, entity screening and validation is the weighted assessment results consisting of the trustworthiness and consistency are the weighted assessment modules, output is the output results of LLMs, and weighted assessment results are the results after screening.

# 3 LLM-DER

The LLM-DER framework is shown in Figure 2. There are three components in total:(1) Relationship List Generation:By predefining specific relationships, entity type information is associated with each relationship to improve the semantic understanding of entity-relationships in LLMs. Subsequently, the LLMs zero-shot learning capability is utilized to generate a list of diversified relationships similar to the predefined relationships to enhance the LLMs perception of entities in the text; (2) Relationship-driven entity recognition:A list of diversified relationships is added to a specific cueing template as a cueing instruction input to the LLMs, and the LLMs utilizes the semantic connection between each diversified relationship and an entity to Determine the entities present in the text and associate the entities; (3) Entity Screening and Validation:After entity recognition, the entity information is distributed among the similar diverse relationships, and the entity slots are associated by employing the designed plausibility and consistency weighted evaluation strategy, and the wrong entities are eliminated. These parts are described in detail below.

## 3.1 Relationship List Generation

Considering that specific domains usually involve multiple entities in different segments, this leads to a situation where entities and relationships are entangled and interact with each other.Therefore, we first predefine context-specific relationships and associate these relationships with the type information of the entities at both ends, with the aim of establishing connections between entities through the relationships and thus enhancing the LLM's perception of the entities in the text. For example, when using "production" as a relationship, the entity types should be "enterprise" and "product" respectively. Based on the fact that the diversity of cueing instructions can further enhance the comprehension ability of LLMs [33,34], we utilize the zero-shot learning ability of LLMs to generate diversified relations that are similar to the predefined relations and associate them with the entity type information respectively. The diversified relations thus generated from each predefined relation are recorded as a list of relations $R_i = \{r_1, r_2 \ldots r_n\}$ , where n denotes the number of diversified relations generated and i denotes the number of predefined relations.

## 3.2 Associated Entity

Based on the generated list of relationships $R_i = \{r_1, r_2 \ldots r_n\}$ , the entities in the text are associated. In addition, we took e ach predefined relation and constructed presentations for LLMs, aiming to fully utilize the ICL capabilities of LLM[35]. Since each relation in the relation list is accompanied by entity type information, LLMs determines the entities present in the text and recognizes the entities by understanding the semantic information of entity-relationships and utilizing the se mantic links between entities and relations. For example, the j-th diversity relation in the i-th relation list results in:$r_j = \{\{subject_1, object_1\}, \{subject_2, object_2\}, \ldots, \{subject_n, object_n\}\}$, where n is the number of matched entity pairs, and we treat the obtained matching result of each diversified relation as a relation pattern candidate for this relation, denoted as $(R_{type}, \{r_1, r_2 \ldots r_n\})$,where $R_{type}$ denotes the predefined relationship and n denotes the number of similar relationships.

## 3.3 Entity Screening And Validation

After associating entities, the entity information is distributed among the relationship schema candidates. Because of the phantom problem of LLMs, this section is to describe how to determine the final entity information by obtaining the two slots from each relationship candidate that are significantly reliable and consistent, i.e., head entity SUBJECT, and tail en tity OBJECT.We use the i-th relationship list $(R_{type}^i, \{r_1^i, r_2^i, \ldots r_n^i\})$,to denote the result of associating entities, where $R_{type}$ denotes the predefined relationship and n denotes the number of similar relationships, where the j-th relationship c andidate is denoted as $r_j^i = (subject1, object1), (subject2, object2), \ldots, (subjectn, objectn)$, We evaluate and associate the ent ity slots by using reliability and consistency metrics, while eliminating the erroneous entities and aggregating the entity i nformation distributed over the relationship candidates to the predefined relationships, and use $R_{type}^i{}' = \{subject_1, object_1\}, \{subject_2, object_2\}, \ldots, \{subject_n, object_n\}$ to denote the result after screening and validated result, where n is t he number of entity pairs, and at this point $R_{type}^i{}'$ is the result of perfect and accurate entity information obtained from all relationship candidates, the pseudo-code of the algorithm is shown below.

| Algorithm1:Relationship-Driven Approach To Entity Identification |
|---|

Input:O:the results of text conceptualization,

    Where i-th instance of O is ($R_{type}^i,\{r_1^i,r_2^i,......r_n^i\}$);

    for i-th instance $\in$ O do

        $R_{type}^i{}' \leftarrow \emptyset$

        for $r_j^i \in \{r_1^i,r_2^i,......r_n^i\}$ do

            $r_j^i$ = "$R_{type}^i$ Slots:$\{subject_1,object_1\},\{subject_2,object_2\},......,\{subject_n,object_n\}$"

            $R_{type}^i{}' \leftarrow R_{type}^i{}' \cup \{subject_1,object_1\},\{subject_2,object_2\},......,\{subject_n,object_n\}$

            for slot $\in R_{type}^i{}'$ do

                Reliability(slot)$^j \leftarrow$ Equation(1)

                Consistency(slot)$^j \leftarrow$ Equation(2)

                Score(slot)j $\leftarrow$ Equation(3)

                if Score(slot)j < threshold do

                $R_{type}^i{}'$.del(slot)

            end for

        end for

        i-th instance $\leftarrow R_{type}^i{}'$

    end for

Return O

Reliability:an entity slot is reliability if it appears frequently with other slots in multiple diverse relationships in a list of relationships, e.g., "The Games of the XXIX Summer Olympiad opened in Beijing, China, the first time that China hosted the Olympic Games", where "Beijing" is reliability because it appears frequently with "China" in the "location" and "capital" diverse relationships. "is reliability because it often occurs together with "China" in "location" and "capital". in both "location" and "capital". In this paper, we use PageRank algorithm [26] to calculate the reliability of entity slots as follows:

$$\text{Reliability(slot)}^j = \beta \sum\nolimits_k^{|R_{type}^i|} \frac{\text{Reliability}(s^k)}{d(s^k)} + (1-\beta)\frac{1}{|R_{type}^i|} \quad (1)$$

where $\beta$ is a hyperparameter,$|R_{type}^j|$ is the number of slots of $R_{type}^j$, $\frac{\text{Reliability}(s^k)}{d(s^k)}$,$(s^k)$denotes that slot s and slot $(s^k)$co-occur in a relation candidate, and the initialized slot reliability score is $\frac{1}{|R_{type}^i|}$.The PageRank algorithm stops after T iterations or when the change in the reliability score is less than a threshold value $\varepsilon$.

Consistency:since LLMs may recognize wrong entities, in order to remove wrong recognition results, we use semantic similarity based on BERT [27] to evaluate the consistency between entities and texts in relation candidates.

$$\text{Consistency(s)j= Sim}(r_{type}^j, T \mid s, r_{type}^j \in c) \qquad (2)$$

where Sim() denotes the semantic similarity function, both s and $r_{type}^j$ come from the candidate relational schema c, and s is an entity slot in the relationship candidate.

The final assessment of a slot by an entity is to have and consistency assessed together[28]:

$$\text{Score(s)i} = (\lambda * \text{Reliability(s)i}) * \text{Consistency(s)i} \qquad (3)$$

Finally, we keep only the entity slots in candidate relationship patterns with the highest confidence scores. Entity slots for candidate relationship patterns with confidence scores below a preset threshold are filtered. By evaluating reliability and consistency together, we effectively mitigate untrue entities identified by the LLMs illusion problem.

## 4 EXPERIMENTAL SETUP

### 4.1 Dataset

Coal: the dataset constructed in this paper in the field of coal chemistry, in which the data are obtained from coal chemical industry websites and some enterprise reports, and manually annotated using the NER annotation tool YEDDA. The entities in this dataset are classified into six categories by the guidance of experts in the field of coal chemical industry: coal chemical products(PRO), industrial policies(POL), downstream industries(IND), upstream raw materials(MAT), process technologies(TEC) and organizations(ORG)(enterprises, groups, research institutes), and Table 2 shows the number of each entity type.

Table2. Statistics on the number of various entity types in the Coal dataset

| Entity type | quantities |
| --- | --- |
| Coal chemical products | 489 |
| Enterprises、Groups、Research institutes | 139 |
| Process technologies | 208 |
| Downstream industries | 232 |
| Upstream raw materials | 224 |
| Industrial policies | 13 |

Resume[16]: a Chinese NER dataset for resumes, which contains eight entity types, including country (CONT), educational background (EDU), place name (LOC), person name (NAME), organization name (ORG), profession (PRO), race (RACE), and title (TITLE).

The specific division of the training, validation and test sets in the above dataset is shown in Table 3.

Table 3.Corpora statistics of the used datasets

| Dataset | Train | Dev | Test | Entity Types |
| --- | --- | --- | --- | --- |
| Resume | 3.8k | 0.46k | 0.48k | 8 |
| Coal | 1.9k | 0.24k | 0.24k | 6 |

## 4.2 Implementation Details

During the training process of fine-tuning the BERT model for similarity computation between sentence and entity pairs, the hyper-parameters were configured as follows: a learning rate of 2e-5, a batch size of 16, and a maximum sequence length of 128 tokens, and 3 cycles of training were performed. The model was implemented and trained using the PyTorch deep learning framework, and the Adam optimizer was used for parameter optimization.

## 4.3 Evaluation

In both experiments on the two datasets in this paper we only consider perfect matching, i.e., we consider a predicted entity to be a correct prediction only if its boundary and type are the same as those of the real entity, and in order to keep up with the baseline model, we similarly evaluate it using accuracy precision(P), recall recall(R), and F1-score(F1), which are defined as follows:

$$P = \frac{N_m}{N_p} \quad (4)$$

$$R = \frac{N_m}{N_r} \quad (5)$$

$$F1 = \frac{2 \times P \times R}{P + R} \quad (6)$$

$N_m$, $N_p$ and $N_r$ denote the total number of correctly predicted entities, predicted entities, and true entities, respectively. F1 is a coordinated average of accuracy and recall, a composite metric that balances the impact of accuracy and recall.

## 4.4 Baseline

In order to evaluate that our proposed LLM-DER framework is more effective for the NER task in the Chinese coal chemical industry, we compare it with the following other methods.

● BERT [30] is a NER method based on BERT that adds a labeled classifier layer downstream of the BERT model.

●Lattice LSTM [16] is a character-based Chinese NER method that uses a lattice-structured LSTM model to introduce dictionary information.

● FLAT [19] is a transformer-based NER method for lattice structures. It constructs a flat-structured transformer to fully utilize the lattice information and take advantage of parallel computation on GPUs.

● PCBERT [29] is a cue-based small-sample NER model for Chinese. It consists of a P-BERT component and a C-BERT component, integrating lexical features and implicit labeling features.

## 5 RESULTS AND ANALYSIS

In this section, we show the experimental results of LLM-DER on the Resume and Coal datasets, as shown in Tables 3 and 4. To evaluate the effectiveness of LLM-DER, we adopt PCBERT's sampling method to simulate few-shot scenarios on Coal and Resume datasets. We took 250, 500, 1000 and 1350 samples from them as training sets to observe the changes of F1 values under different data ranges, and these values were used as evaluation metrics for the final performance.

Table 4.Experimental results on the Resume dataset

| Model | Resume NER | | | |
| --- | --- | --- | --- | --- |
| | K=250 | K=500 | K=1000 | K=1350 |
| BERT | 53.80 | 62.64 | 69.36 | 70.65 |
| Lattice LSTM | 85.63 | 89.60 | 92.01 | 93.13 |
| FLAT | 84.62 | 90.77 | 92.97 | 87.79 |
| PCBERT | 93.42 | 94.01 | 94.96 | 95.97 |
| Ours | 92.64 | 93.21 | 93.58 | 93.94 |

Table 5. Experimental results of coal chemical dataset

| | Coal NER | | | |
|---|---|---|---|---|
| Model | K=250 | K=500 | K=1000 | K=1350 |
| BERT | 50.73 | 55.27 | 56.74 | 56.76 |
| Lattice-LSTM | 65.42 | 66.34 | 67.24 | 67.54 |
| FLAT | 65.35 | 66.18 | 66.93 | 67.48 |
| PCBERT | 66.28 | 67.38 | 68.32 | 68.46 |
| Ours | 67.91 | 68.13 | 69.22 | 70.53 |

## 5.1 Main Results

According to the results in Tables 4 and 5, our method did not demonstrate significant advantages over the chosen SOTA model on the public dataset Resume. However, on the coal chemical dataset, our method significantly outperforms the chosen model and achieves the highest F1 value in every sample size condition. This suggests that our method has an advantage in domain-specific entity recognition tasks.In addition, we also observe that there is a significant difference in the performance of our method on the Resume and Coal datasets. On the Coal dataset, as the number of samples K increases from 250 to 1350, we observe changes in F1 values of 0.22%, 1.09%, and 1.31%, respectively; while on the Resume dataset, these changes are 0.57%, 0.37%, and 0.36%, respectively. This suggests that our approach exhibits higher sensitivity and adaptability when dealing with the coal-chemical domain with complex entity structure patterns, leading to larger fluctuations in F1 values with increasing sample size. In contrast, the texts in the resume dataset typically lack complex entity structures, and thus we observe smaller variations in the F1 values. This phenomenon highlights the performance of our approach in processing domain-specific data in terms of its feature sensitivity and adaptability to data complexity.

Thus, the method of generating a list of relations by predefining specific relations increases the diversity and enables LLMs to understand complex structures and recognize entities more accurately. The weighted assessment of reliability and consistency removes possible recognition errors due to the LLMs illusion problem, thus enhancing entity recognition. To further validate the effectiveness of LLM-DER on the Coal dataset, we performed ablation analysis.

## 5.2 Ablation studies on the Coal dataset

To validate the effectiveness of the relationship list module in LLM-DER and the weighted consistency assessment module for avoiding the LLMs illusion problem, we conducted ablation experiments on the Coal dataset. Table 6 shows in detail the impact of each module on the performance of the named entity recognition task.

Table 6 . Coal dataset ablation results

| | Coal NER | | | |
|---|---|---|---|---|
| Model | | | | |
| | K=250 | K=500 | K=1000 | K=1350 |

| | | | |
|---|---|---|---|
| LLM-DER | 67.91 | 68.13 | 69.22 | 70.53 |
| w/o relationship list | 65.24 | 66.14 | 67.24 | 68.21 |
| w/o estimate | 66.04 | 67.32 | 68.16 | 69.41 |

In Table 6, we draw the following conclusions: (1) The relationship list module significantly enhances the ability of LLMs to understand complex structures by predefining relationships in a specific context and correlating them with the type information of the entities at both ends, which in turn improves the accuracy of entity recognition, and additionally, generating diverse relationships related to the type of the target entity ensures that more target entities can be correctly recognized, which in turn improves the recall; (2) the removal of the entity evaluation module leads to a performance degradation, and although the diverse relationships enhance the ability to understand complex structures, they may also increase the interference with the LLMs; (3) the case in which the F1 value for the removal of the evaluation module is higher than that for the removal of the relationship list module is due to the fact that the relationship list can help the LLMs to capture the complex structure of the domain in a more comprehensive way. Although the weighted assessment of reliability and consistency of entity slots can effectively remove erroneous entities, thus improving the overall accuracy and recall. However, the effectiveness of this weighted assessment strategy presupposes that LLMs can comprehensively capture entities in complex structures.

## 6 CONCLUSION

This study aims to solve the domain-specific NER problem and takes the coal chemical industry domain as a study case for domain-specific entity recognition. First, to address the scarcity of domain-specific labeled data and the insufficiency of few-shot learning methods, we apply GPT-3.5-Turbo to the domain NER task and design the LLM-DER framework, which employs two strategies to recognize relevant entities:(1) adding entity type information at both ends of predefined relationships, and using contextual relationships to establish connections between entities, thus enhancing the LLMs perception of entities in the text (2) designing reliability and consistency weighted assessment methods to remove misrecognized entities. The experimental results on Resume dataset and Coal dataset show that LLM-DER has excellent performance in the coal chemical industry.

Since the training data of the large language model is from the general domain and there is no data from the specialized domain, to better apply the LLMs to the domain-specific entity recognition task, in the future, we can consider trying to fine-tune the current large language model, such as LLaMA2, using domain-specific corpus to further optimize the task performance.